\documentclass[sigconf]{acmart}
\usepackage{multirow}
\usepackage{balance}

\AtBeginDocument{%
  \providecommand\BibTeX{{%
    \normalfont B\kern-0.5em{\scshape i\kern-0.25em b}\kern-0.8em\TeX}}}

\copyrightyear{2020}
\acmYear{2020}
\setcopyright{acmcopyright}\acmConference[MM '20]{Proceedings of the 28th ACM International Conference on Multimedia}{October 12--16, 2020}{Seattle, WA, USA}
\acmBooktitle{Proceedings of the 28th ACM International Conference on Multimedia (MM '20), October 12--16, 2020, Seattle, WA, USA}
\acmPrice{15.00}
\acmDOI{10.1145/3394171.3414053}
\acmISBN{978-1-4503-7988-5/20/10}

\newcommand\blfootnote[1]{%
\begingroup 
\renewcommand\thefootnote{}\footnote{#1}%
\addtocounter{footnote}{-1}%
\endgroup 
}

\acmSubmissionID{mmfp1555}
\begin{document}
\fancyhead{}

\title{Fine-grained Iterative Attention Network for Temporal Language Localization in Videos}


\author{Xiaoye Qu$^\dagger$}
\affiliation{%
  \institution{School of Electronic Information and Communication, Huazhong University of Science and Technology 
  \& Huawei Tech., Hangzhou, China}}
\email{xiaoye@hust.edu.cn}

\author{Pengwei Tang$^\dagger$}
\affiliation{%
  \institution{School of Electronic Information and Communication, Huazhong University of Science and Technology}}
\email{pengweitang@hust.edu.cn}

\author{Zhikang Zou}
\affiliation{%
  \institution{Department of Computer Vision Technology (VIS), Baidu Inc., China}}
\email{zouzhikang@baidu.com}

\author{Yu Cheng}
\affiliation{%
  \institution{Microsoft Dynamics 365 AI Research}
  \city{Redmond, WA}
  \country{United States of America}}
\email{yu.cheng@microsoft.com}

\author{Jianfeng Dong}
\affiliation{%
  \institution{School of Computer and Information Engineering, Zhejiang Gongshang University}}
\email{dongjf24@gmail.com}

\author{Pan Zhou$^*$}
\affiliation{%
  \institution{The Hubei Engineering Research Center on Big Data Security, School of Cyber Science and Engineering, Huazhong University of Science and Technology}}
\email{panzhou@hust.edu.cn}

\author{Zichuan Xu}
\affiliation{%
  \institution{School of Software, Dalian University of Technology}}
\email{z.xu@dlut.edu.cn}


\begin{abstract}

Temporal language localization in videos aims to ground one video segment in an untrimmed video based on a given sentence query. To tackle this task, designing an effective model to extract ground-ing information from both visual and textual modalities is crucial. However, most previous attempts in this field only focus on unidirectional interactions from video to query, which emphasizes which words to listen and attends to sentence information via vanilla soft attention, but clues from query-by-video interactions implying where to look are not taken into consideration. In this paper, we propose a Fine-grained Iterative Attention Network (FIAN) that consists of an iterative attention module for bilateral query-video in-formation extraction. Specifically, in the iterative attention module, each word in the query is first enhanced by attending to each frame in the video through fine-grained attention, then video iteratively attends to the integrated query. Finally, both video and query information is utilized to provide robust cross-modal representation for further moment localization. In addition, to better predict the target segment, we propose a content-oriented localization strategy instead of applying recent anchor-based localization. We evaluate the proposed method on three challenging public benchmarks: Ac-tivityNet Captions, TACoS, and Charades-STA. FIAN significantly outperforms the state-of-the-art approaches.

\end{abstract}





\maketitle

\blfootnote{$^\dagger$Equal Contribution.}
\blfootnote{$^*$Corresponding author: Pan Zhou.}

\section{Introduction}
\begin{figure}
\setlength{\belowcaptionskip}{-1.5em}
\centering
\includegraphics[width=8.5cm]{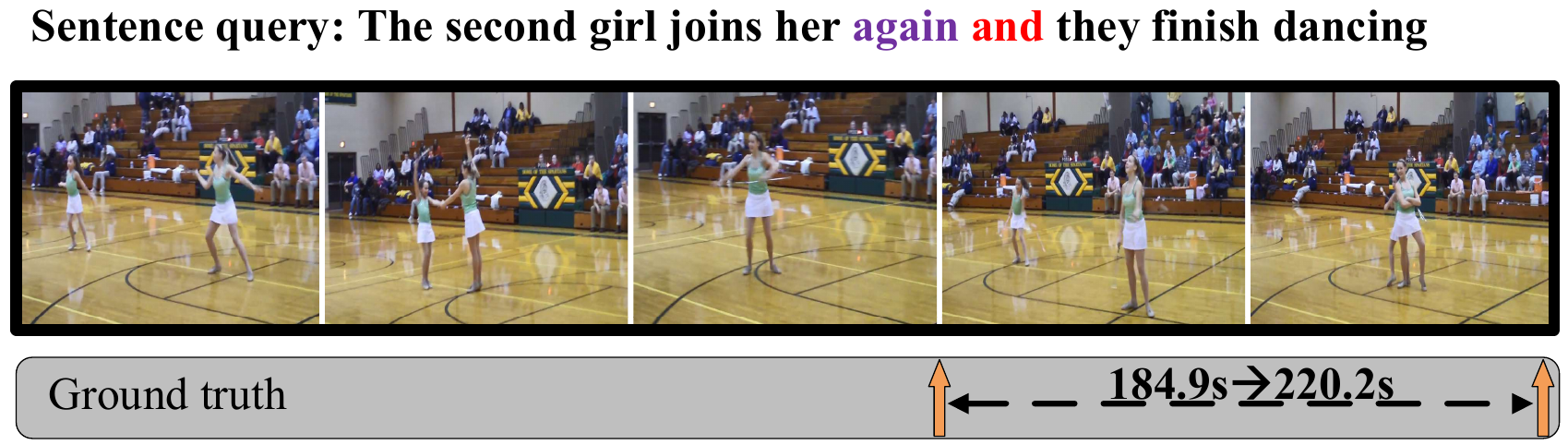}
\caption{Temporal language localization is designed to localize a video segment with a start point (184.9s) and an end point (220.2s) in an untrimmed video corresponding to the given sentence query.} 
\label{Figure-query}
\end{figure}

Localizing activities is a challenging yet pragmatic task for video understanding. In real scenarios, a video may contain multiple activities of interests which are associated with complex language dependencies, and cannot be classified to a pre-defined list of action classes. To solve this problem, temporal language localization \cite{Gao2017TALLTA} is proposed and attracts increasing attention recently. Formally, as shown in Figure \ref{Figure-query}, given an untrimmed video and a sentence query, this task is to automatically identify the start and end boundaries of the video segment semantically corresponding to the given sentence query. Comparing with other video researches, such as video retrieval and video captioning, this task is more challenging as fine-grained interactions between video and sentence need to be modelled to differentiate video segments in the long video. 

Considering the query ``The second girl joins her again and they finish dancing" depicted in Figure \ref{Figure-query}, which emphasizes that ``the second girl" appears with a temporal relation ``again". A model that only localizes the action ``The second girl joins her" is not satisfactory, since this action appears twice in the video. Therefore, designing an effective model to collect grounding information from both modalities is central to task performance. From the video perspective, it is necessary to capture detailed temporal contents and decide which part does the sentence describe. From the sentence aspect, as several words or phrases can give clear cues to identify the target video segment, we should pay more attention to these sentence details.
It is conceivable that attending to key words with the video and highlighting critical frames with query both contribute to precise location.


To solve this task, traditional methods \cite{Gao2017TALLTA,liu2018attentive,ge2019mac} for temporal language localization first sample candidate video segments using sliding windows, and then fuse the sentence with each video segment representations separately to calculate the matching relationships. However, they integrate global sentence representations with video segment representations via matrix operations rather than explore the fine-grained interactions across video and sentence. 
Recently, some work \cite{chen-etal-2018-temporally,Chen2019LocalizingNL,Zhang2019CrossModalIN} integrate the whole video with sentence query to generate a sentence-aware video representation for further location prediction. Specifically, they aggregate word features in the sentence for each frame with the widely used soft attention to obtain distinguishable frame features.
While promising results have been achieved by these works, they fail to adequately exploit the sentence semantic information, since they merely explore the unidirectional interaction from video to the sentence.
In this paper, we propose a novel Fine-grained Iterative Attention Network (FIAN), which iteratively performs attention for multi-modal location information gathering. The main idea is as straightforward as to collect grounding clues from both modalities. In specific, we design a cross-modal guided attention (CGA) to capture the fine-grained interactions from different modalities in multiple feature spaces. With CGA, each component from two modalities achieves comprehensively interaction. Furthermore, 
to integrate the attended information from CGA, we propose a cross-modal encoder to iteratively generate sentence-aware video and video-aware sentence representations. These two representations are then incorporated into the cross-modal feature space with filter controlling. To the end, we are able to obtain a robust cross-modal feature for subsequent temporal localization.

Besides, in order to fully utilize the cross-modal information for temporal localization, we devise a content-oriented location strategy. Different from traditional anchor-based prediction which simultaneously predicts multi-scale windows by feature at each time step, we utilize the complete features inside each window for prediction. In this way, each candidate window can be comprehensively evaluated, thus leading to precise localization.
Overall, the main contributions of this work are: 
\begin{itemize}
\item We propose a novel Fine-grained Iterative Attention Network (FIAN) for temporal language localization, in which fine-grained sentence and video grounding information is attended. To our best knowledge, we are the first work to explicitly utilize both video-aware sentence and sentence-aware video representations for accurate location.

\item We devise a content-oriented localization strategy to better predict the temporal boundary. Based on the cross-modal information, it carefully measures the whole component in each candidate window for candidate moment evaluation. 

\item We conduct experiments on three public datasets: ActivityNet Captions, TACoS, and Charades-STA and FIAN significantly outperform the state-of-the-art by a large margin.
\end{itemize}

\begin{figure*}
\centering
\includegraphics[width=17cm]{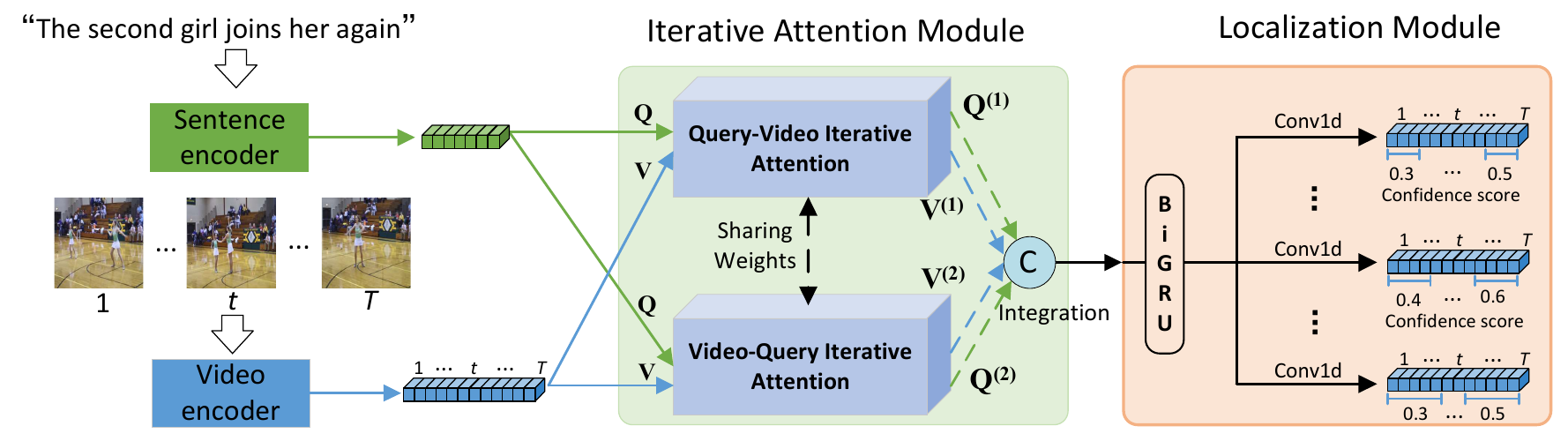}
\caption{An overview of our Fine-grained Iterative Attention Network (FIAN) for temporal language localization which consists of three parts: (1) Video and sentence query are encoded into feature representations. (2) Based on the two modal information, symmetrical iterative attentions generate both video-aware sentence and sentence-aware video representations for two attention branches. Then all representations are integrated to produce the cross-modal information. (3) The localization module finally locates the boundaries of target moments.}
\label{SEN}
\end{figure*}

\section{Related works}\label{sec:rw}
\subsection{Temporal Action Localization}
Temporal action localization aims to locate action instances in an untrimmed video. Approaches for this task can be classified into three categories: (1) methods performing frame or segment-level classification where the smoothing and merging steps are substantially required to obtain the temporal boundaries \cite{shou2017cdc,zeng2019breaking}. (2) methods adopting a two-stage framework involving proposal generation, classification and boundary refinement \cite{videvent,shou2016temporal,xu2017r,zhao2017temporal}. (3) methods developing a end-to-end architecture integrating the proposal generation and classification \cite{Wang2016CVPR,yeung2016end,lin2017single}. 
Although these works have achieved promising performance, they are limited to a pre-defined list of actions. Thus, temporal language localization is proposed to tackle this issue by introducing the language query.

\subsection{Language Localization in images}
Language localization in images is also called ``locating referring expressions in images", which aims to localize the object instance in an image described by a referring expression phrased in natural language. Traditional works in this field solve this task using a CNN/LSTM framework \cite{hu2016natural,mao2016generation,nagaraja2016modeling}. The LSTM takes as input a region-level CNN feature and a word vector at each
time step, and aims to maximize the likelihood of the expression given the referred region. Another line of work treats referring expression comprehension as a metric learning problem \cite{luo2017comprehension,rohrbach2016grounding,wang2016learning}, whereby the expression feature and the region feature are embedded into a common feature space to measure the compatibility. The focus of these approaches lies in how
to define the matching loss function. These approaches tend to use a single feature vector to represent the expression and the image region. To overcome this limitation of monolithic features, self-attention mechanisms have been used to decompose the
expression into sub-components and learn separate features for each of the resulting parts \cite{hu2017modeling,yu2018mattnet,zhang2018grounding}.

\subsection{Temporal language localization in videos}
Temporal language localization in videos requires understandings of both complex video scenes and natural language, which is a new task introduced recently \cite{Gao2017TALLTA,Hendricks2017LocalizingMI}. Several early methods \cite{liu2018attentive,ge2019mac} sample video segments through dense sliding windows. After aggregating the two modality information using simple matrix operation, they measure the similarity between these candidate video segments and sentence query in a common embedding space. In this way, the temporal language localization degrades into a multi-modal matching problem. While simple and effective, these methods just consider global representation and fail to exploit the fine-grained interaction between modalities. 

Recently, in order to avoid the redundant computation by predefined sliding windows, some works propose to integrate sentence query information with the whole video first, and then predict the temporal boundary by directly regressing the start and end points \cite{yuan2019find,Chen2019LocalizingNL,Lu2019DEBUGAD} or designing multi-scale temporal anchors \cite{Zhang_2019_CVPR,Zhang2019CrossModalIN} which follow the same spirit of anchor box in object detection. These methods closely integrate the video and sentence representations and obtain improved performance. They usually adopt frame-by-word interactions for language and video feature fusion, which aggregates sentence information for each frame according to the normalized similarity.   
However, these methods lack fully exploring the sentence semantic information which plays an important role in distinguishing the ambiguous frames and strengthening the integrated video representations for precise localization.
Our proposed FIAN captures the more fine-grained interaction between two modalities, thus leading to more distinguished features. Moreover, our work firstly utilizes the video-aware sentence representation for enhancing the sentence-aware video representation, in order to obtain a robust cross-modal information for following moment prediction.

\section{Model Description}\label{sec:model}
In this section, we first introduce the basic formation of temporal language localization. Then we present the detailed structure of our fine-grained iterative attention network, which consists of feature representation, symmetrical iterative attention module, and moment localization module. The whole structure of our network is shown in Figure \ref{SEN}.

\subsection{Problem Formulation}
Given an untrimmed video $V$, and a natural language query $Q$, the task aims to localize the temporal video segment described by the query. We represent the video frame-by-frame as ${V}=\left\{{v}_{i}\right\}_{i=1}^{n_v}$, where $v_i$ is the feature of $i$-th frame of the video and $n_v$ is the frame number of the video. Similarly, the given natural language query can be denoted as ${Q}=\left\{{q}_{i}\right\}_{i=1}^{n_q}$ word-by-word, where $q_i$ is the feature of the corresponding word. Our goal is to predict the start and end temporal coordinates $(s,e)$ in the video. 

\subsection{Representation For Language and Video}
 \textbf{Query representation} For query encoding, we obtain its embedding vector by the Glove \cite{Pennington2014GloveGV}. To aggregate the contextual information, we then feed the sentence embedding to the bi-directional GRU network \cite{chung2014empirical}. Formally, given word features $Q=(q_1,q_2,...,q_{n_q})$, we obtain the contextual representation of each word by:  
 %
\begin{align}
h_i^f=&{\text{GRU}_q^f}(q_i,h_{(i-1)}^f)\\
h_i^b=&{\text{GRU}_q^b}(q_i,h_{(i-1)}^b)\\
h_i^q=&[h_i^f;h_i^b]
\end{align}     
where ${\text{GRU}_q^f}$ and ${\text{GRU}_q^b}$ are the forward and backward GRU network. Thus, we get the contextual query representations $h_q=(h_1^q,h_2^q,...,h^q_{n_q})$ by concatenating the forward and backward hidden state at each time step. \\
\textbf{Video representation} Similar to \cite{Zhang2019CrossModalIN}, for video information encoding, we first extract features from the pre-trained 3D network and then employ the self-attention \cite{Vaswani2017AttentionIA} to learn the semantic dependencies in the long video context.
 Also, in order to learn the contextual information within the video, we feed the video features into the bi-directional GRU network to further incorporate the contextual information. In the similar way, we can get the video representation $h_v=(h_1^v,h_2^v,...,h^v_{n_v})$.
 
\begin{figure}
\setlength{\belowcaptionskip}{-1.5em}
\centering
\includegraphics[width=8.5cm]{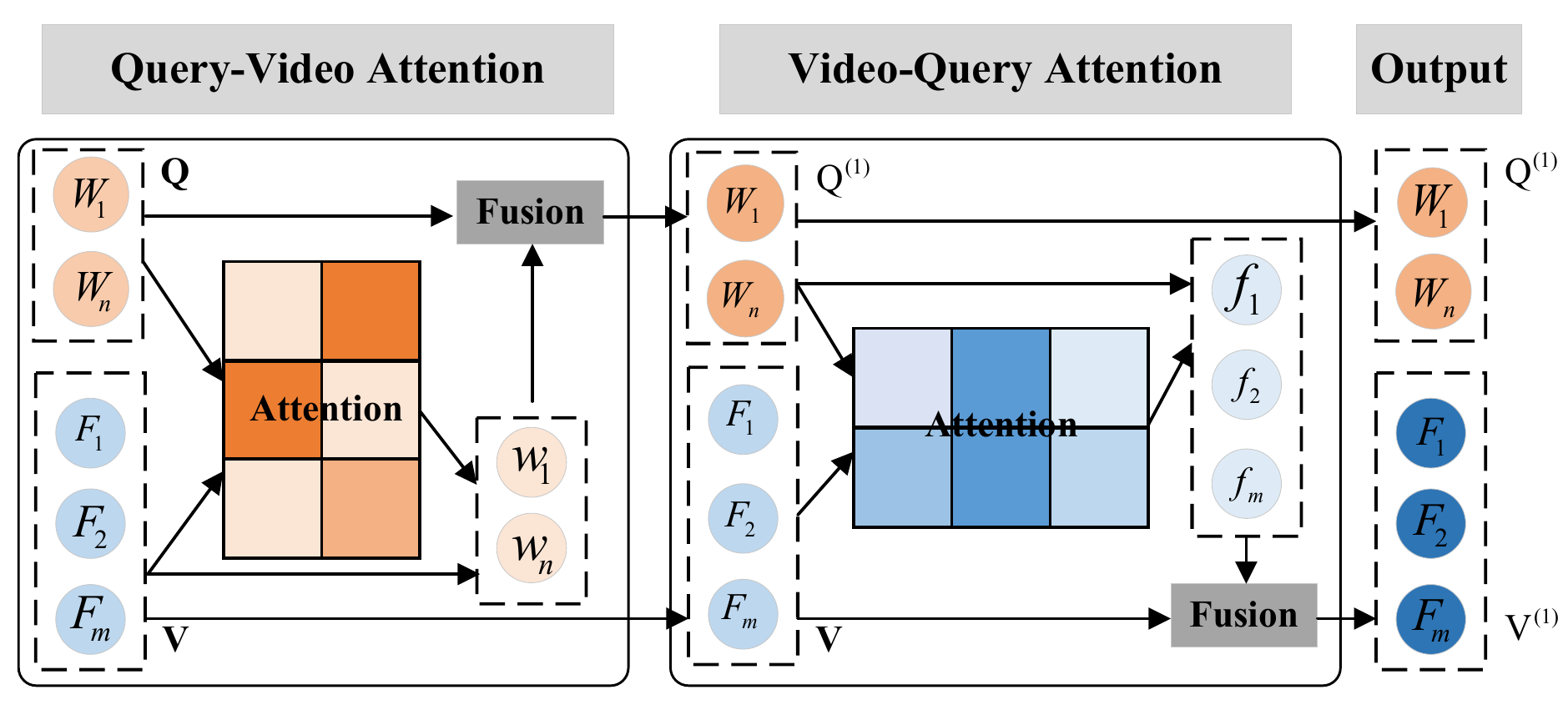}
\caption{The structure of query-video iterative attention. The query first attends to video to obtain integrated query information and then the video iteratively attends to the integrated query for further enhanced video. During each attention mechanism, each word in the query and each frame in the video achieve fine-grained interactions.} 
\label{Figure-VQ}
\end{figure}

\subsection{Iterative Attention Module} 
After obtaining the video and sentence representation, our aim is to achieve fine-grained cross-modal interaction and collect grounding information from both modalities. To this end, as shown in Figure \ref{SEN}, we design an symmetrical iterative attention module which contains two sub-modules sharing same architecture but reverse input, namely query-video iterative attention and video-query iterative attention. As in Figure \ref{Figure-VQ}, the query-video iterative attention successively performs query-video attention and video-query attention to obtain integrated query and video information. Next, we will describe the most important components in this architecture including attention mechanism and information fusion.

\begin{figure}
\centering
\includegraphics[width=8.5cm]{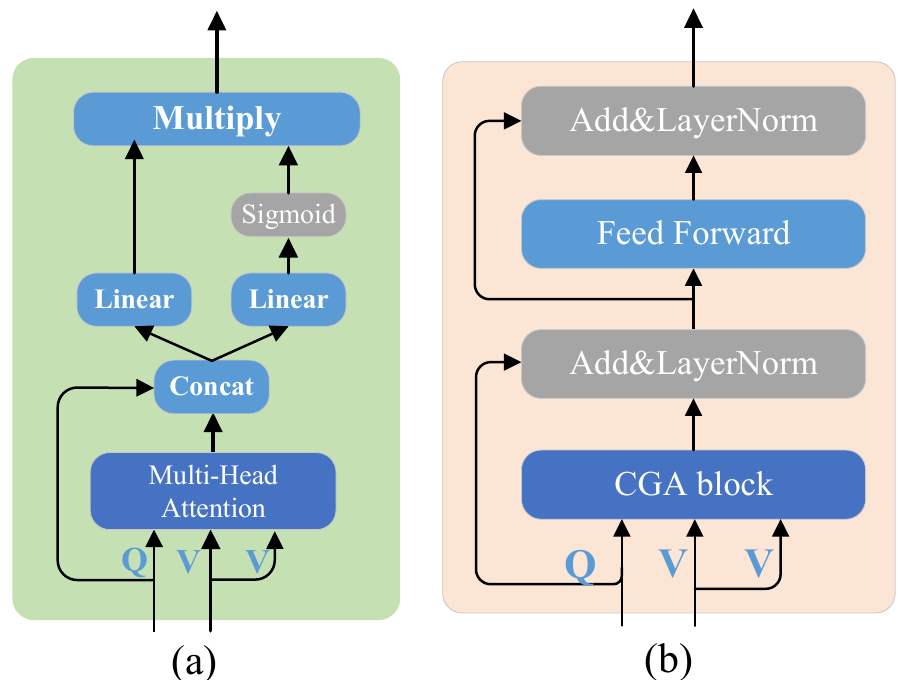}
\caption{\textbf{Left:} The structure of cross-modal guided attention (CGA). \textbf{Right:} The architecture of \textbf{Cross-Modal Encoder}. We name this encoder as a Q-V encoder which generates V-aware Q representation. }
\label{Figure-CGA}
\end{figure} 

\subsubsection{Cross-Modal Guided Attention}
\ \\
In our localization task, a robust model needs to find out the exact starting and ending point of the video segment, however, a simple soft attention mechanism which captures interaction from one specific attention space may not be enough to solve this problem. Thus, we devise a cross-modal guided attention (CGA) consisting of cross-modal multi-head attention (CMA) and information gate for this task, as shown in Figure \ref{Figure-CGA} (a).

Here we first introduce CMA whose inputs are query $Q \in \mathbb{R}^{n_q \times d_{h}}$, and value ${V} \in \mathbb{R}^{n_{v}\times{d_h}}$ from two different modalities, where $n_{q}, n_{v}$ represent the numbers of queries and values and $d_{h}$ represents feature dimension. The multi-head attention is composed of $n$ parallel heads and each head performs the scaled dot-product attention as:
%
\begin{equation}
\text{Att}_{i}(Q,V)=\text{Softmax}\left(\frac{Q W_{i}^{Q}\left(V W_{i}^{K}\right)^{\top}}{\sqrt{d_{k}}}\right) V W_{i}^{V}
\end{equation}%
where ${W_{i}^Q},{W_{i}^K},{W_{i}^V}\in\mathbb{R}^{d_{h}\times d_{k}}$ are learnable parameter matrices of projections. $d_k = d_h/n$ is the size of the output features for each head.
The outputs yielded by each head are concatenated and then projected again to construct the final output:
\begin{equation}
\text{ CMA }(Q, V) = \text{ Concat (head}_{1},\ldots,\text{head}_{n}) W^{O}
\end{equation}%
where $\text{head}_{i}=\text{Att}_{i}(Q, V)$, $W^O\in\mathbb{R}^{d_{h}\times d_{h}}$ is the linear parameter.

With the help of CMA, we can capture the cross-modal interactions from multiple different aspects corresponding to various attention heads. Furthermore, to obtain more distinguishing features, we employ an information gate \cite{huang2019attention} on CMA to refine the output information as noisy information could be captured by partial heads. In this process, the outputs from CMA are filtered according to the query semantic meaning, namely the attended information is guided by query information. We name this guided CMA as CGA and the CGA is computed as:
\begin{align}
\text{CGA}(Q,V) =& \sigma (W_{q}^{g} Q+W_{v}^{g}\text{CMA}({Q},{V})+b^{g})\nonumber\\
\odot& (W_{q}^{i}{Q}+W_{v}^{i}\text{ CMA }({Q}, {V})+b^{i}) 
\end{align}%
where $W_{q}^{g}, W_{v}^{g}, W_{q}^{i}, W_{v}^{i}\in\mathbb{R}^{d_h\times d_h}$, $b^{g},b^{i}\in\mathbb{R}^{d_h}$, and $\odot$ denotes element-wise multiplication. In practice, we keep $W_{q}^{g} =W_{q}^{i}$ and $W_{v}^{g} =W_{v}^{i}$ by sharing projection weights at training stage. After the CGA, we can achieve effective interactions between two modalities.

\subsubsection{Information Fusion}
\ \\
After CGA, we attempt to fuse query $Q$ with attended information $\text{CGA}(Q,V)$ to enhance the the query representation. Inspired by Transformer's encoder, we apply a cross-modal encoder structure in Figure \ref{Figure-CGA} (b) to integrate these two branches. Here we just present the feed-forward procedure by:  
\begin{align}
\mathrm{FFN}(X)=\max (0, X W^{(1)}+b^{(1)}) W^{(2)}+b^{(2)}
\end{align}
but omit the internal shortcut connection \cite{he2016deep}, and layer normalization \cite{ba2016layer}. $W^{(1)}$ and $W^{(2)}$ are linear transformation parameters, $b^{(1)}$ and $b^{(2)}$ are bias parameters.  
Thus, the cross-modal encoder can be denoted as:%
\begin{align}
\text{Encoder} (Q,V)= \text{FFN}(\text{CGA}(Q,V))
\end{align}%
This encoder intrinsically captures the fine-grained interaction from query to value as it computes the similarity for each query element to all value elements, and generates a \textbf{V-aware Q representation}. We call this encoder as \textbf{Q-V encoder}. Similarly, the V-Q encoder can yield a Q-aware V representation.

Based on above cross-modal encoder, the query-video iterative attention in Figure 3 can be denoted as:
\begin{align}
{{Q}^{(1)}}= \text{Encoder}({Q}, {V}), \;
{{V}^{(1)}}= \text{Encoder}({V}, {Q}^{(1)})
\end{align}%
Analogously, the video-query iterative attention in Figure 2 can be presented as:
%
\begin{align}
{{V}^{(2)}}= \text{Encoder}({V}, {Q}), \;
{{Q}^{(2)}}= \text{Encoder}({Q}, {{V}^{(2)}})
\end{align}%
These two iterative attention branch compensate each other and contributes to generating robust cross-modal information. 

\subsubsection{Video-enhanced Integration}
\ \\
With the symmetrical iterative attention module, we obtain two sentence-aware video representations ${V}^{(1)}$, ${V}^{(2)} \in \mathbb{R}^{n_v \times d_{h}}$ and two video-aware query representations ${Q}^{(1)}$, ${Q}^{(2)} \in \mathbb{R}^{n_q \times d_{h}}$. To complement the sentence-aware video representation for further enhanced video information, we first project each query representation to same length as video by:%
\begin{align}
\hat{{Q}}^{(1)}= {W}_{1}{{Q}^{(1)}}+b_{1},\;\hat{{Q}}^{(2)}= {W}_{2}{{Q}^{(2)}}+b_{2} 
\end{align}%
where ${W}_{1}\in \mathbb{R}^{n_v \times n_{q}}$ and ${W}_{2}\in \mathbb{R}^{n_v \times n_{q}}$ are matrices for linear transformation, $b_{1}\in \mathbb{R}^{n_{q}}$ and $b_{2}\in \mathbb{R}^{n_{q}}$ are the bias terms. We then combine the corresponding modality information by column as: %
\begin{align}
\hat{V}=\text{Concat}[{{V}}^{(1)},{{V}}^{(2)}], \; \hat{Q}=\text{Concat}[\hat{Q}^{(1)},{\hat{Q}}^{(2)}]\;
\end{align}%

Finally, we devise a filter to control the ratio of query information to incorporate with the video features. 
%
\begin{align}
\boldsymbol{r}&=\sigma(\hat{Q}W^{r}+b^{r})
\\
{M}&=\text{LayerNorm}(\hat{{V}}+\boldsymbol{r}\odot\hat{{Q}})
\end{align}%
where $W^r \in \mathbb{R}^{2d_h \times 2d_{h}}$, $b^r \in \mathbb{R}^{2d_h}$ are learnable parameters and $\odot$ denotes element-wise multiplication.

\subsection{Localization Module}
In this section, we introduce the localization module. Previous anchor-based predictions \cite{chen-etal-2018-temporally,Zhang2019CrossModalIN} simultaneously score a set of candidate moments with multi-scale windows at each time step, however, the confidence scores of different window scales are predicted based on the same point feature. In this paper, we devise a content-oriented strategy which differs from traditional anchor-based localization. During our localization process, the confidence scores of candidate moments with multi-scale windows can be calculated with the entire features in the corresponding time duration.

\begin{figure}
\includegraphics[width=8.5cm]{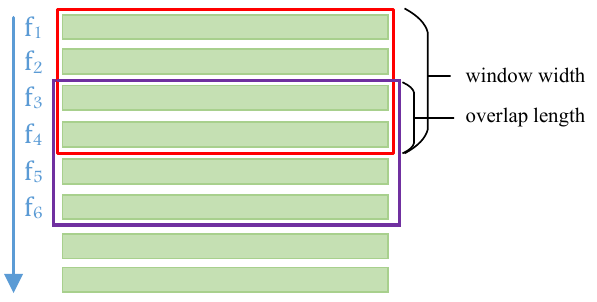}
\caption{
Red and pink rectangles denote representation sequence corresponding to candidate moments which can be marked as $C_1=(1,d), C_2=(1+d-p, 2d-p)$, where $d$ and $p$ represent window width and overlap length. More candidate moments can be sampled by overlapped sliding windows in the same way.  
}
\label{fig:prediction}
\end{figure}

With the integrated cross-modal representation ${M}$, we first apply a bi-directional GRU network to aggregate contextual information, resulting in final representation sequence $\hat{M}=(f_1,f_2,\cdots,f_{n_v})$. To predict the target video segment, we pre-define several candidate moments for grounding by dividing representation sequence into overlapped windows. 
In practice, we have multi-size windows. Taking one window size for example, as shown in Figure \ref{fig:prediction}, $j$-th candidate moment can be denoted as $C_j=(\hat{s}_j, \hat{e}_j)$, where $\hat{s}_j, \hat{e}_j$ are the starting and ending coordinates between 1 to $n_v$.  
Subsequently, our goal is to score these candidate moments and adjust the temporal boundary for them. 
Here we adopt temporal 1D convolution to process features of each candidate moment to produce the corresponding confidence score ${cs_j}$ and temporal offsets $\hat\delta_{j} = (\hat{{\delta}^{s}}, \hat{{\delta}^{e}})$.   
The temporal 1D convolution can be simply denoted as $\text{Conv1d}(C_f,\theta_{k},\theta_{s})$, where $C_f,\theta_{k},\theta_{s}$ are filter numbers, kernel size and stride size, respectively. 
Given a representation sequence corresponding to candidate moment $C_j$, we apply two distinct convolution $\text{Conv1d}(1,\theta_{k},\theta_{s})$ and $\text{Conv1d}(2,\theta_{k},\theta_{s})$ for producing confidence score and temporal offsets separately. Then the confidence scores will be normalized by sigmoid function.  
$\theta_{k}$ is same as the window size and $\theta_{s}$ is equal to window size minus overlap length. In this way, temporal 1D convolution can properly process all candidate moments.

\subsection{Training Loss and Inference}
We first compute the IoU (Intersection over Union) $o_j$ between each candidate moment $(\hat{s}_j, \hat{e}_j)$ with ground truth $(s,e)$. If $o_j$ is larger than a threshold value $\tau$, this candidate moment is viewed as positive sample, reverse as the negative sample. Thus we can obtain $N_{pos}$ positive samples and $N_{neg}$ negative samples in total.
\\
\textbf{Alignment Loss}: We adopt an alignment loss to align the predicted confidence scores $cs$ and IoU $o$, which promotes candidate moments with higher IoUs achieve higher confidence scores, the alignment loss is calculated as:
\begin{align}
&\mathcal{L}_{j} =o_{j}\log \left(cs_{j}\right)+\left(1-o_{j}\right) \log \left(1-cs_{j}\right)\\
&\mathcal{L}_{align}=\sum_{z\in\{pos,neg\}}-\frac{1}{N_{z}}\sum_{j}^{N_{z}} \mathcal{L}_{j}
\end{align}%
\textbf{Boundary Loss}: As the boundaries of pre-defined candidate moments are relatively coarse, we devise a boundary loss for $N_{pos}$ positive samples to promote exploring the precise start and end points. The boundary loss is:
\begin{align}
\mathcal{L}_{b}=\frac{1}{N_{pos}} \sum_{j}^{N_{pos}} \mathcal{R}_{1}\left(\hat\delta_{j}^{s}-\delta_{j}^{s}\right)+ \mathcal{R}_{1}\left(\hat\delta_{j}^{e}-\delta_{j}^{e}\right)
\end{align}%
where $\mathcal{R}_1$ represents the smooth L1 function, $\delta_{j}^{s}$ and $\delta_{j}^{e}$ are the starting and ending offsets of the ground-truth coordinates compared to the pre-defined candidate coordinates, and $\hat\delta_{j}^{s}$ and  $\hat\delta_{j}^{e}$ are the predicted offsets. 
\\\textbf{Joint Loss}: We adopt $\alpha$ to control the balance of the alignment loss and boundary loss:
\begin{align}
&\mathcal{L}=\mathcal{L}_{align}+\alpha \mathcal{L}_{b}
\end{align}%
\textbf{Inference}: We rank all candidate moments according to their predicted confidence scores, and then ``Top-n (Rank@n)" candidates will be selected with non maximum suppression.

\section{Experiments}

\subsection{Datasets}
We validate our proposed approach on three datasets. \\
\textbf{ActivityNet Captions}~\cite{Krishna2017DenseCaptioningEI}: It is a large dataset which contains 20k videos with 100k language descriptions. The video contents
of this dataset are diverse and open. This dataset pays attention to complicated human activities in daily life. Following public split, we use 37,417, 17,505, and 17,031 sentence-video pairs for training, validation, and testing respectively. \\
\textbf{TACoS}~\cite{Regneri2013GroundingAD}: It collects 127 long  videos, which are mainly about cooking scenarios. On this dataset, we use the same split as ~\cite{Gao2017TALLTA}, which has 10146, 4589 and 4083 sentence-video pairs for training, validation, and testing.\\
\textbf{Charades-STA}~\cite{Gao2017TALLTA}: Gao et al.\cite{Gao2017TALLTA} first label the start and end time of moments) of this dataset with language descriptions. It consists of 9,848 videos of daily life indoors activities. There are 12,408 sentence-video pairs for training and 3,720 pairs for testing.\\
\subsection{Evaluation Metrics}
Following previous works~\cite{Zhang2019CrossModalIN,SCDM19}, we adopt ``Rank @n, IoU @m" as evaluation metrics. ``Rank @n, IoU @m" is defined as the percentage of the language queries having at least one matched retrieval (IoU with ground-truth moment is larger than m) in the top-n retrieved moments. 

\subsection{Implementation Details}
Following~\cite{SCDM19}, we adopt C3D~\cite{Tran2014LearningSF} for ActivityNet Captions and TACoS, and I3D~\cite{Carreira2017QuoVA} for Charades-STA to encode videos. Next, a fully-connected layer is used to reduce video feature dimension to 512. We set the length of video feature sequences to 200 for ActivityNet Captions and TACoS, and 64 for Charades-STA. For too long videos, we downsample them uniformly. During prediction, we use convolution kernel size of [16, 32, 64, 96, 128, 160, 192] for ActivityNet Captions, [8, 16, 32, 64] for TACoS, and [16, 24, 32, 40] for Charades-STA. We then set stride size as 0.25, 0.125, 0.125 of kernel size for ActivityNet Captions, TACoS and Charades-STA, respectively. The trade-off parameter $\alpha$ is set 0.001 for ActivityNet Captions, 0.005 for TACoS and Charades-STA. In symmetrical mutual attention, we set heads as 8 for ActivityNet Captions and TACoS, and 4 for Charades-STA. The positive threshold value $\tau$ is set to 0.55. We train our model using Adam optimizer ~\cite{Kingma2014AdamAM} with learning rate of ${8} \times {10}^{-4}$, ${4} \times {10}^{-4}$, and ${4} \times {10}^{-4}$ for ActivityNet Captions, TACoS, and Charades-STA, respectively. The batch size is set to 128, 64, and 64, respectively. Hidden dimension of all bi-directional GRUs is set as 512 in our model.

\subsection{Performance Comparison}
We compare our FIAN with existing state-of-the-art methods, which can be classified into: (1) Sliding window-based models:  
MCN~\cite{Hendricks2017LocalizingMI}, CTRL~\cite{Gao2017TALLTA}, ACRN~\cite{liu2018attentive}, QSPN~\cite{xu2019multilevel}, TripNet~\cite{Hahn2019TrippingTT}. (2) Recent works generate sentence-aware video representations: ABLR~\cite{yuan2019find}, MAN~\cite{Zhang_2019_CVPR}, 
CMIN~\cite{Zhang2019CrossModalIN}, SCDM~\cite{SCDM19}. The performance comparisons of previous methods on three public benchmarks are shown in Table 1-3. We can observe that the FIAN achieves a new state-of-the-art performance under nearly all evaluation metrics and benchmarks. 

\begin{table}[t!]
    \small
    \centering
    \caption{Performance compared with previous methods on the Activity Captions dataset.}
    \label{tab:compare1}
    \setlength{\tabcolsep}{1.5mm}{
    \begin{tabular}{c|cccccc}
    \hline \hline
    \multirow{2}*{Method} & R@1 & R@1 & R@1 & R@5 & R@5 & R@5 \\ 
    ~ & IoU=0.3 & IoU=0.5 & IoU=0.7 & IoU=0.3 & IoU=0.5 & IoU=0.7 \\ \hline
    MCN \cite{Hendricks2017LocalizingMI} & 39.35 & 21.36 & 6.43  & 68.12 & 53.23 & 29.70\\
    CTRL \cite{Gao2017TALLTA} & 47.43 & 29.01 & 10.34 & 75.32 & 59.17 & 37.54\\
    QSPN \cite{xu2019multilevel} & 45.30 & 27.70 & 13.60 & 75.70 & 59.20 & 38.30\\
    TripNet \cite{Hahn2019TrippingTT} & 48.42 & 32.19 & 13.93&- & -&-\\
    ACRN \cite{liu2018attentive} & 49.70 & 31.67 & 11.25 & 76.50 & 60.34 & 38.57\\
    ABLR \cite{yuan2019find} & 55.67 & 36.79 &   -   &  -   &   -   &  - \\
    CMIN \cite{Zhang2019CrossModalIN} & 63.61 & 43.40 & 23.88 & 80.54 & 67.95 & 50.73\\
    SCDM \cite{SCDM19} & 54.80 & 36.75 & 19.86 & 77.29 & 64.99 & 41.53\\ \hline
    \textbf{FIAN} & \textbf{64.10} & \textbf{47.90} & \textbf{29.81} & \textbf{87.59} & \textbf{77.64} & \textbf{59.66} \\ \hline
    \end{tabular}}
\end{table}

\begin{table}[t!]
    \small
    \centering
    \caption{Performance compared with previous methods on the TACoS dataset.}
    \label{tab:compare1}
    \setlength{\tabcolsep}{1.5mm}{
    \begin{tabular}{c|cccccc}
    \hline \hline
    \multirow{2}*{Method} & R@1 & R@1 & R@1 & R@5 & R@5 & R@5 \\ 
    ~ & IoU=0.1 & IoU=0.3 & IoU=0.5 & IoU=0.1 & IoU=0.3 & IoU=0.5 \\ \hline
    MCN \cite{Hendricks2017LocalizingMI} & 14.42 & - & 5.58 & 37.35 & - & 10.33 \\ 
    CTRL \cite{Gao2017TALLTA} & 24.32 & 18.32 & 13.30 & 48.73 & 36.69 &25.42\\
    QSPN \cite{xu2019multilevel} & 25.31 & 20.15 & 15.23 & 53.21 & 36.72 & 25.30\\
    ABLR \cite{yuan2019find} & 34.70 & 19.50 &   9.40   &  -   &   -   &  - \\
    TripNet \cite{Hahn2019TrippingTT} & - & 23.95 & 19.17& -&-&-\\
    ACRN \cite{liu2018attentive} & 24.22 & 19.52 & 14.62 & 47.42 & 34.97 & 24.8\\
    CMIN \cite{Zhang2019CrossModalIN} & 32.48 & 24.64 & 18.05 & \textbf{62.13} & 38.46 & 27.02\\
    SCDM \cite{SCDM19} & - & 26.11 & 21.17 & - & 40.16 & 32.18\\ \hline
    \textbf{FIAN} & \textbf{39.55} & \textbf{33.87} & \textbf{28.58} & \textbf{56.14} & \textbf{47.76} & \textbf{39.16} \\ \hline
    \end{tabular}}
\end{table}

\begin{table}[t!]
    \small
    \centering
    \caption{Performance compared with previous methods on the Charades-STA dataset.}
    \label{tab:compare1}
    \setlength{\tabcolsep}{2mm}{
    \begin{tabular}{c|cccccc}
    \hline \hline
    \multirow{2}*{Method} & R@1 & R@1 & R@5 & R@5 \\ 
    ~ & IoU=0.5 & IoU=0.7 & IoU=0.5 & IoU=0.7 \\ \hline
    MCN \cite{Hendricks2017LocalizingMI} & 17.46 & 8.01 & 48.22 & 26.73 \\
    CTRL \cite{Gao2017TALLTA} & 23.63 & 8.89 & 58.92 & 29.52\\
    QSPN \cite{xu2019multilevel} & 35.60 & 15.80 & 79.40 & 45.40\\
    TripNet \cite{Hahn2019TrippingTT} & 36.61 & 14.50 & -& -\\
    ACRN \cite{liu2018attentive} & 20.26 & 7.64 & 71.99 & 27.79 \\
    MAN \cite{Zhang_2019_CVPR} & 46.53 & 22.72 & 86.23 & 53.72\\
    SCDM \cite{SCDM19} & 54.44 & 33.43 & 74.43 & 58.08\\ \hline
    \textbf{FIAN} & \textbf{58.55} & \textbf{37.72} & \textbf{87.80} & \textbf{63.52}\\ \hline
    \end{tabular}}
\end{table}

\noindent\textbf{ActivityNet Captions.} As we can see from Table 1, FIAN brings 5.93\% improvement in the strict ``R@1, IoU=0.7" metric, and outperforms around 10\% in the all R@5 metrics than the previous state-of-the-art method in absolute values.

\begin{table}[t!]
    \small
    \centering
    \caption{Ablation study on the TACoS dataset.}
    \label{tab:compare1}
    \setlength{\tabcolsep}{1mm}{
    \begin{tabular}{c|cccccc}
    \hline \hline
    \multirow{2}*{Method} & R@1 & R@1 & R@1 & R@5 & R@5 & R@5 \\ 
    ~ & IoU=0.1 & IoU=0.3 & IoU=0.5 & IoU=0.1 & IoU=0.3 & IoU=0.5 \\ \hline
    FIAN-Soft & 34.97 & 27.53 & 21.64 & 52.27 & 43.99 & 32.52 \\
    FIAN-CMA & 36.82 & 29.27 & 23.05 & 53.86 & 44.40 & 34.61\\
    FIAN-VQ & 36.32 & 29.32 & 24.03 & 53.56 & 44.72 & 37.08 \\
    FIAN-VQMA & 36.91 & 30.59 & 25.54 &54.37& 45.28 & 37.30\\
    \hline
    FIAN-QVMA & 36.59 & 30.10 & 24.81 & 53.15 & 43.84 & 35.42  \\
    \hline
    FIAN-Concat &38.03&31.59&26.01&54.86&46.14&36.71\\
    FIAN-Matrix& 38.31 & 32.21 &  25.96   &  55.45  &   45.43 &  37.50 \\
    \hline
    \textbf{FIAN} & \textbf{39.55} & \textbf{33.87} & \textbf{28.58} &  \textbf{56.14} & \textbf{47.76} & \textbf{39.16}\\ \hline
    \end{tabular}}
\end{table}

\noindent\textbf{TACoS.} Table 2 shows that our proposed FIAN 
achieves around 7\% higher improvements than previous methods in all metrics except the ``Rank@5, IoU=0.1". Compared to other datasets, the performances of TACoS are worst, which may come from similar background and objects during the whole video. However, it is worth noting that FIAN still achieves significant improvements, which demonstrates that our proposed model can effectively differentiate the visual similarity frames.\\
\textbf{Charades-STA}. In Table 3, we can observe that FIAN surpasses much over SCDM, especially 13.37\% in ``Rank5, IoU=0.5" and 5.44\% in ``Rank 5, IoU=0.7", demonstrating that FIAN has powerful capacity of generating robust cross-modal information for prediction.\\
\textbf{Overall Analysis}. 
Compared to state-of-the-art methods, FIAN gains significant improvements, especially in the stiff ``Rank1, IoU=0.5" or ``Rank1, IoU=0.7" metrics. 
The results of sliding window-based methods are obviously inferior to recent methods. ABLR directly predicts start and end points based on the features rather than design windows for prediction, thus leading to inaccurate results.
MAN and CMIN only exploit sentence-aware video representation, while SCDM further devises a mechanism to dynamically modulate the sentence-aware video representation. Although these methods achieve promising performance, they do not explore the video-aware sentence representation to enhance the cross-modal information. From the results, we can see that FIAN achieves better results by utilizing both sentence-aware video and video-aware sentence representations.

\begin{figure*}
\centering
\includegraphics[width=16cm]{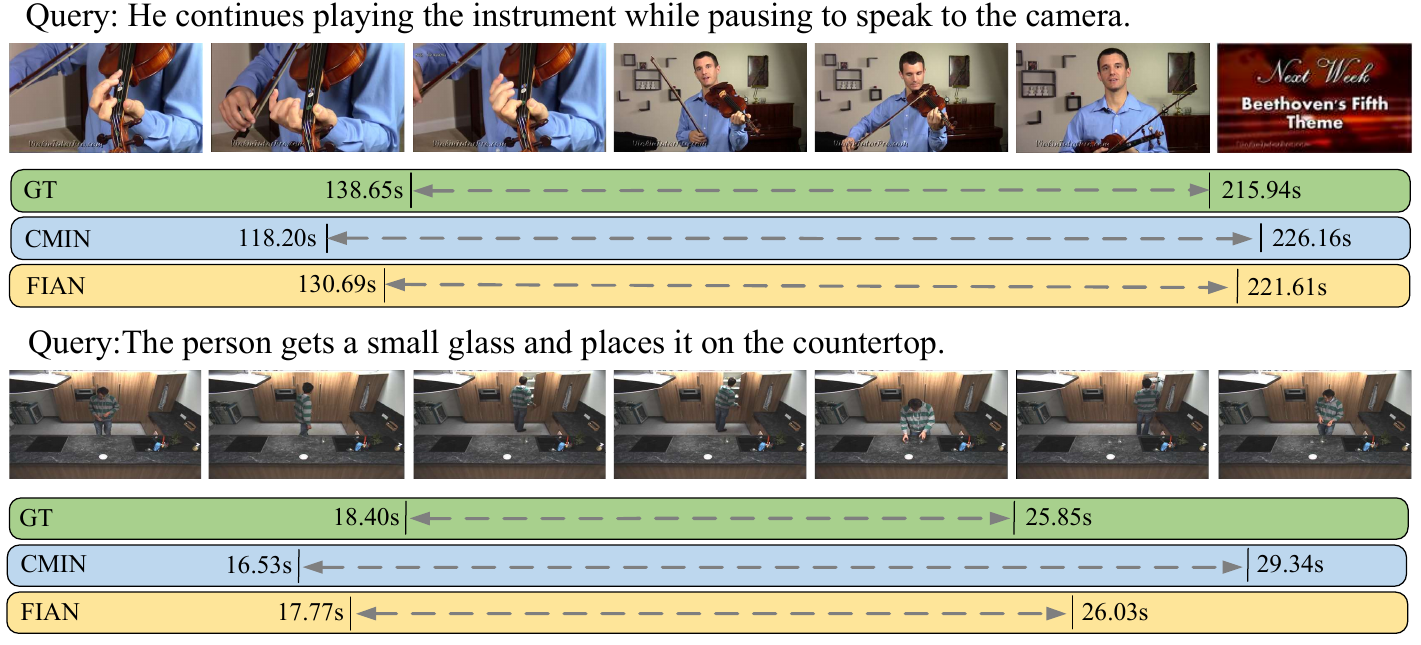}
\caption{Qualitative visualization of temporal language localization results (Rank@1) by CMIN, and FIAN. First example comes from Activity Captions and second example is from TACoS dataset. Ground truth (GT) is also provided for both samples.
}
\label{fig:model5}
\end{figure*}

\begin{table}[H]
    \small
    \centering
    \caption{Ablation study on the TACoS dataset. We apply different localization strategy to substitute the localization module in our network. FIAN-Full is our method with content-oriented location strategy.}
    \label{tab:compare1}
    \setlength{\tabcolsep}{1mm}{
    \begin{tabular}{c|cccccc}
    \hline \hline
    \multirow{2}*{Method} & R@1 & R@1 & R@1 & R@5 & R@5 & R@5 \\ 
    ~ & IoU=0.1 & IoU=0.3 & IoU=0.5 & IoU=0.1 & IoU=0.3 & IoU=0.5 \\ 
    \hline
    FIAN-Full&39.55&33.87&28.58&56.14&47.76&39.16\\
    \hline
    FIAN-TGN&34.82&23.37&20.05&48.76&38.40&31.64\\
    \hline
    FIAN-CMIN&35.56&27.46&22.39&51.41&39.92&33.88\\
    \hline
    \end{tabular}}
\end{table}

\begin{table}[H]
    \small
    \centering
    \caption{Ablation study on the TACoS dataset. We evaluate the influence of different stride sizes in the localization module. $\theta_k$ is the kernel size.}
    \label{tab:compare1}
    \setlength{\tabcolsep}{1mm}{
    \begin{tabular}{c|cccccc}
    \hline \hline
    \multirow{2}*{Stride Size} & R@1 & R@1 & R@1 & R@5 & R@5 & R@5 \\ 
    ~ & IoU=0.1 & IoU=0.3 & IoU=0.5 & IoU=0.1 & IoU=0.3 & IoU=0.5 \\ 
    \hline
    1&37.45&29.83&21.09&51.68&40.73&29.90\\
    \hline
    1/8$\theta_{k}$ & \textbf{39.55} & \textbf{33.87} & \textbf{28.58} &  56.14 & 47.76 & 39.16\\
    \hline
    1/4$\theta_{k}$&37.01&30.32&25.84&57.09&45.09&37.89\\
    \hline
    1/2$\theta_{k}$&38.43&31.40&25.37&\textbf{60.72}&\textbf{51.14}&\textbf{41.81}\\
    \hline
    \end{tabular}}
\end{table}

\section{Ablation Study}

\subsection{Influence of fine-grained iterative attention}
In this section, we present ablation studies to understand the effects of CGA block, video-aware sentence representation, and symmetrical attention. We re-train our approach with following settings: 

\begin{itemize}
    \item \textbf{FIAN-Soft}: Instead of symmetrical iterative attention, we just use one video-query encoder and substitute the CGA in the encoder with soft attention. 
    \item \textbf{FIAN-CMA}: We adopt one video-query encoder as above while substituting the CGA with CMA.
    \item \textbf{FIAN-VQ}: Compared with above, we apply one complete video-query encoder to generate sentence-aware video representation $V^{(2)}$ as shown in Figure 2.
    \item \textbf{FIAN-VQMA}: We utilize one iterative attention branch to generate sentence-aware video representation $V^{(2)}$ and video-aware sentence representation $Q^{(2)}$.     
    \item \textbf{FIAN-QVMA}: We apply iterative attention to generate sentence-aware video representation $V^{(1)}$ and corresponding sentence representation $Q^{(1)}$.
    \item \textbf{FIAN-Concat}: We directly concatenate two video representations $[{{V}}^{{(1)}},{{V}}^{{(2)}}]$ from two iterative attention branches instead of applying video-enhanced integration.
\item \textbf{FIAN-Matrix}: We use $[{{V}}^{(1)},{{V}}^{(2)},{{V}}^{(1)}-{{V}}^{(2)},{V}^{(1)}\odot{{V}}^{(2)}]$ to fuse the video representations from two branches instead of fusion gate.
    \item \textbf{FIAN}: Our full of FIAN model.
\end{itemize}


Table 4 shows the performance comparisons of our FIAN and these ablations on the most difficult TACoS dataset. \\
\textbf{Effect of CGA block.} Comparing FIAN-CMA with FIAN-Soft, it is significant that multi-head attention obtains a more fine-grained interaction than plain soft attention. With information gate on FIAN-CMA, FIAN-VQ effectively performs filtering information for all candidates, which is demonstrated in ``Rank@5, IoU=0.5". Thus, our CGA block captures more detailed cross-modal interactions than widely used soft attention. \\  
\textbf{Effect of video-aware sentence representation.} (1) Comparing FIAN-VQMA with FIAN-VQ, it can be observed that ``Rank@1, IoU=0.5" improves 1.51\% and performance gains in all metrics. It indicates that video-aware sentence representation can also benefit cross-modal information integration for prediction. 
(2) Comparing FIAN-Concat and FIAN-Matrix with FIAN, direct concatenation or performing matrix operations between video representations show weaker results. It denotes that video-aware sentence information from two branches also effectively enhances the cross-modal integration. \\
\textbf{Effect of symmetrical iterative attention.} Comparing FIAN-QVMA and FIAN-VQMA with FIAN, single branch 
performs worse than full FIAN, which demonstrates that two branches from symmetrical mutual attention can compensate each other to generate more robust cross-modal information.  \\



\subsection{Influence of location module}
In this section, we first compare our localization module with existing techniques by replacement in order to further verify the effectiveness of the proposal. Here we replace our localization module with corresponding component in TGN \cite{chen-etal-2018-temporally}, and CMIN \cite{Zhang2019CrossModalIN}, and rename these variants as FIAN-TGN and FIAN-CMIN. The performance degeneration observed in Table 5 verifies the superiority of our proposed modules over their competitors. Moreover, we adjust different stride sizes $\theta_s$ during candidate moments sampling. The results are shown in Table 6. We can observe that too many candidate moments (stride=1) leads to performance drop. This indicates that dense candidate moments confuse the learning process of regression. Meanwhile, too few candidate moments (stride=1/2$\theta_k$) brings the highest R@5 metrics. For more precise R@1 retrieving results, we choose 1/8$\theta_k$ as our experiments setting.

\subsection{Qualitative Results}
To qualitatively validate the locating effectiveness of our FIAN, we show two examples on ActivityNet Captions and TACoS dataset. 
As shown in Figure \ref{fig:model5}, in both samples, FIAN is capable of locating an event which consists of two diverse activities, although the target moment in ActivityNet Captions is obviously longer than the one in TACoS dataset. By intuitive comparison, our FIAN localizes more accurate boundaries than CMIN \cite{Zhang2019CrossModalIN} and explicitly decreases the location error brought by ambiguous frames. 

\section{Conclusions}
In this paper, we propose a novel Fine-grained Iterative Attention Network (FIAN) to adequately extract bilateral video-query interaction information for temporal language localization in videos. Besides proposing a refined cross-modal guided attention (CGA) block to capture the detailed cross-modal interactions, FIAN further adopts a symmetrical iterative attention to generate both sentence-aware video and video-aware sentence representations, where the latter is explicitly facilitated to enhance the former and finally both parts contribute to a robust cross-modal feature. In addition, we devise a content-oriented localization strategy to better predict the temporal boundary. Extensive experiments on three real-world datasets validate the effectiveness of our method. 

The future work includes apply FIAN to more complicated benchmarks \cite{liu2020violin,lei2018tvqa}. Combining our method with some pre-training model \cite{li2020hero} could also further boost the performance. 

\section{Acknowledgments}

This work was supported in part by the National Natural Science Foundation of China (No. 61972448), National Natural Science Foundation of China (No. 61902347), and Zhejiang Provincial Natural Science Foundation (No. LQ19F020002).


\balance
\bibliographystyle{ACM-Reference-Format}
\bibliography{sample-base}


\begin{thebibliography}{45}


\ifx \showCODEN    \undefined \def \showCODEN     #1{\unskip}     \fi
\ifx \showDOI      \undefined \def \showDOI       #1{#1}\fi
\ifx \showISBNx    \undefined \def \showISBNx     #1{\unskip}     \fi
\ifx \showISBNxiii \undefined \def \showISBNxiii  #1{\unskip}     \fi
\ifx \showISSN     \undefined \def \showISSN      #1{\unskip}     \fi
\ifx \showLCCN     \undefined \def \showLCCN      #1{\unskip}     \fi
\ifx \shownote     \undefined \def \shownote      #1{#1}          \fi
\ifx \showarticletitle \undefined \def \showarticletitle #1{#1}   \fi
\ifx \showURL      \undefined \def \showURL       {\relax}        \fi
\providecommand\bibfield[2]{#2}
\providecommand\bibinfo[2]{#2}
\providecommand\natexlab[1]{#1}
\providecommand\showeprint[2][]{arXiv:#2}

\bibitem[\protect\citeauthoryear{Ba, Kiros, and Hinton}{Ba
  et~al\mbox{.}}{2016}]%
        {ba2016layer}
\bibfield{author}{\bibinfo{person}{Jimmy~Lei Ba}, \bibinfo{person}{Jamie~Ryan
  Kiros}, {and} \bibinfo{person}{Geoffrey~E Hinton}.}
  \bibinfo{year}{2016}\natexlab{}.
\newblock \showarticletitle{Layer normalization}.
\newblock \bibinfo{journal}{\emph{arXiv preprint arXiv:1607.06450}}
  (\bibinfo{year}{2016}).
\newblock


\bibitem[\protect\citeauthoryear{Carreira and Zisserman}{Carreira and
  Zisserman}{2017}]%
        {Carreira2017QuoVA}
\bibfield{author}{\bibinfo{person}{Jo{\~a}o Carreira} {and}
  \bibinfo{person}{Andrew Zisserman}.} \bibinfo{year}{2017}\natexlab{}.
\newblock \showarticletitle{Quo Vadis, Action Recognition? A New Model and the
  Kinetics Dataset}. In \bibinfo{booktitle}{\emph{Proceedings of the IEEE
  Conference on Computer Vision and Pattern Recognition}}.
  \bibinfo{pages}{4724--4733}.
\newblock


\bibitem[\protect\citeauthoryear{Chen, Chen, Ma, Jie, and Chua}{Chen
  et~al\mbox{.}}{2018}]%
        {chen-etal-2018-temporally}
\bibfield{author}{\bibinfo{person}{Jingyuan Chen}, \bibinfo{person}{Xinpeng
  Chen}, \bibinfo{person}{Lin Ma}, \bibinfo{person}{Zequn Jie}, {and}
  \bibinfo{person}{Tat-Seng Chua}.} \bibinfo{year}{2018}\natexlab{}.
\newblock \showarticletitle{Temporally Grounding Natural Sentence in Video}. In
  \bibinfo{booktitle}{\emph{Proceedings of the Conference on Empirical Methods
  in Natural Language Processing}}.
\newblock


\bibitem[\protect\citeauthoryear{Chen, Ma, Chen, Jie, and Luo}{Chen
  et~al\mbox{.}}{2019}]%
        {Chen2019LocalizingNL}
\bibfield{author}{\bibinfo{person}{Jingyuan Chen}, \bibinfo{person}{Lin Ma},
  \bibinfo{person}{Xinpeng Chen}, \bibinfo{person}{Zequn Jie}, {and}
  \bibinfo{person}{Jiebo Luo}.} \bibinfo{year}{2019}\natexlab{}.
\newblock \showarticletitle{Localizing Natural Language in Videos}. In
  \bibinfo{booktitle}{\emph{Proceedings of the AAAI Conference on Artificial
  Intelligence}}.
\newblock


\bibitem[\protect\citeauthoryear{Cheng, Fan, Pankanti, and Choudhary}{Cheng
  et~al\mbox{.}}{2014}]%
        {videvent}
\bibfield{author}{\bibinfo{person}{Yu Cheng}, \bibinfo{person}{Quanfu Fan},
  \bibinfo{person}{Sharath Pankanti}, {and} \bibinfo{person}{Alok Choudhary}.}
  \bibinfo{year}{2014}\natexlab{}.
\newblock \showarticletitle{Temporal Sequence Modeling for Video Event
  Detection}. In \bibinfo{booktitle}{\emph{CVPR}}.
\newblock


\bibitem[\protect\citeauthoryear{Chung, Gulcehre, Cho, and Bengio}{Chung
  et~al\mbox{.}}{2014}]%
        {chung2014empirical}
\bibfield{author}{\bibinfo{person}{Junyoung Chung}, \bibinfo{person}{Caglar
  Gulcehre}, \bibinfo{person}{KyungHyun Cho}, {and} \bibinfo{person}{Yoshua
  Bengio}.} \bibinfo{year}{2014}\natexlab{}.
\newblock \showarticletitle{Empirical evaluation of gated recurrent neural
  networks on sequence modeling}.
\newblock \bibinfo{journal}{\emph{arXiv preprint arXiv:1412.3555}}
  (\bibinfo{year}{2014}).
\newblock


\bibitem[\protect\citeauthoryear{Gao, Sun, Yang, and Nevatia}{Gao
  et~al\mbox{.}}{2017}]%
        {Gao2017TALLTA}
\bibfield{author}{\bibinfo{person}{Jiyang Gao}, \bibinfo{person}{Chen Sun},
  \bibinfo{person}{Zhenheng Yang}, {and} \bibinfo{person}{Ramakant Nevatia}.}
  \bibinfo{year}{2017}\natexlab{}.
\newblock \showarticletitle{TALL: Temporal Activity Localization via Language
  Query}. In \bibinfo{booktitle}{\emph{Proceedings of the IEEE International
  Conference on Computer Vision (ICCV)}}. \bibinfo{pages}{5277--5285}.
\newblock


\bibitem[\protect\citeauthoryear{Ge, Gao, Chen, and Nevatia}{Ge
  et~al\mbox{.}}{2019}]%
        {ge2019mac}
\bibfield{author}{\bibinfo{person}{Runzhou Ge}, \bibinfo{person}{Jiyang Gao},
  \bibinfo{person}{Kan Chen}, {and} \bibinfo{person}{Ram Nevatia}.}
  \bibinfo{year}{2019}\natexlab{}.
\newblock \showarticletitle{MAC: Mining Activity Concepts for Language-based
  Temporal Localization}. In \bibinfo{booktitle}{\emph{2019 IEEE Winter
  Conference on Applications of Computer Vision (WACV)}}. IEEE,
  \bibinfo{pages}{245--253}.
\newblock


\bibitem[\protect\citeauthoryear{Hahn, Kadav, Rehg, and Graf}{Hahn
  et~al\mbox{.}}{2019}]%
        {Hahn2019TrippingTT}
\bibfield{author}{\bibinfo{person}{Meera Hahn}, \bibinfo{person}{Asim Kadav},
  \bibinfo{person}{James~M. Rehg}, {and} \bibinfo{person}{Hans~Peter Graf}.}
  \bibinfo{year}{2019}\natexlab{}.
\newblock \showarticletitle{Tripping through time: Efficient Localization of
  Activities in Videos}. In \bibinfo{booktitle}{\emph{Proceedings of the IEEE
  Conference on Computer Vision and Pattern Recognition,Workshop}}.
\newblock


\bibitem[\protect\citeauthoryear{He, Zhang, Ren, and Sun}{He
  et~al\mbox{.}}{2016}]%
        {he2016deep}
\bibfield{author}{\bibinfo{person}{Kaiming He}, \bibinfo{person}{Xiangyu
  Zhang}, \bibinfo{person}{Shaoqing Ren}, {and} \bibinfo{person}{Jian Sun}.}
  \bibinfo{year}{2016}\natexlab{}.
\newblock \showarticletitle{Deep residual learning for image recognition}. In
  \bibinfo{booktitle}{\emph{Proceedings of the IEEE conference on computer
  vision and pattern recognition}}. \bibinfo{pages}{770--778}.
\newblock


\bibitem[\protect\citeauthoryear{Hendricks, Wang, Shechtman, Sivic, Darrell,
  and Russell}{Hendricks et~al\mbox{.}}{2017}]%
        {Hendricks2017LocalizingMI}
\bibfield{author}{\bibinfo{person}{Lisa~Anne Hendricks},
  \bibinfo{person}{Oliver Wang}, \bibinfo{person}{Eli Shechtman},
  \bibinfo{person}{Josef Sivic}, \bibinfo{person}{Trevor Darrell}, {and}
  \bibinfo{person}{Bryan~C. Russell}.} \bibinfo{year}{2017}\natexlab{}.
\newblock \showarticletitle{Localizing Moments in Video with Natural Language}.
  In \bibinfo{booktitle}{\emph{Proceedings of the IEEE International Conference
  on Computer Vision (ICCV)}}. \bibinfo{pages}{5804--5813}.
\newblock


\bibitem[\protect\citeauthoryear{Hu, Rohrbach, Andreas, Darrell, and Saenko}{Hu
  et~al\mbox{.}}{2017}]%
        {hu2017modeling}
\bibfield{author}{\bibinfo{person}{Ronghang Hu}, \bibinfo{person}{Marcus
  Rohrbach}, \bibinfo{person}{Jacob Andreas}, \bibinfo{person}{Trevor Darrell},
  {and} \bibinfo{person}{Kate Saenko}.} \bibinfo{year}{2017}\natexlab{}.
\newblock \showarticletitle{Modeling relationships in referential expressions
  with compositional modular networks}. In
  \bibinfo{booktitle}{\emph{Proceedings of the IEEE Conference on Computer
  Vision and Pattern Recognition}}. \bibinfo{pages}{1115--1124}.
\newblock


\bibitem[\protect\citeauthoryear{Hu, Xu, Rohrbach, Feng, Saenko, and
  Darrell}{Hu et~al\mbox{.}}{2016}]%
        {hu2016natural}
\bibfield{author}{\bibinfo{person}{Ronghang Hu}, \bibinfo{person}{Huazhe Xu},
  \bibinfo{person}{Marcus Rohrbach}, \bibinfo{person}{Jiashi Feng},
  \bibinfo{person}{Kate Saenko}, {and} \bibinfo{person}{Trevor Darrell}.}
  \bibinfo{year}{2016}\natexlab{}.
\newblock \showarticletitle{Natural language object retrieval}. In
  \bibinfo{booktitle}{\emph{Proceedings of the IEEE Conference on Computer
  Vision and Pattern Recognition}}. \bibinfo{pages}{4555--4564}.
\newblock


\bibitem[\protect\citeauthoryear{Huang, Wang, Chen, and Wei}{Huang
  et~al\mbox{.}}{2019}]%
        {huang2019attention}
\bibfield{author}{\bibinfo{person}{Lun Huang}, \bibinfo{person}{Wenmin Wang},
  \bibinfo{person}{Jie Chen}, {and} \bibinfo{person}{Xiao-Yong Wei}.}
  \bibinfo{year}{2019}\natexlab{}.
\newblock \showarticletitle{Attention on Attention for Image Captioning}. In
  \bibinfo{booktitle}{\emph{Proceedings of the IEEE International Conference on
  Computer Vision (ICCV)}}.
\newblock


\bibitem[\protect\citeauthoryear{Kingma and Ba}{Kingma and Ba}{2014}]%
        {Kingma2014AdamAM}
\bibfield{author}{\bibinfo{person}{Diederik~P. Kingma} {and}
  \bibinfo{person}{Jimmy Ba}.} \bibinfo{year}{2014}\natexlab{}.
\newblock \showarticletitle{Adam: A Method for Stochastic Optimization}.
\newblock \bibinfo{journal}{\emph{CoRR}}  \bibinfo{volume}{abs/1412.6980}
  (\bibinfo{year}{2014}).
\newblock


\bibitem[\protect\citeauthoryear{Krishna, Hata, Ren, Li, and Niebles}{Krishna
  et~al\mbox{.}}{2017}]%
        {Krishna2017DenseCaptioningEI}
\bibfield{author}{\bibinfo{person}{Ranjay Krishna}, \bibinfo{person}{Kenji
  Hata}, \bibinfo{person}{Frederic Ren}, \bibinfo{person}{Fei-Fei Li}, {and}
  \bibinfo{person}{Juan~Carlos Niebles}.} \bibinfo{year}{2017}\natexlab{}.
\newblock \showarticletitle{Dense-Captioning Events in Videos}. In
  \bibinfo{booktitle}{\emph{Proceedings of the IEEE International Conference on
  Computer Vision (ICCV)}}. \bibinfo{pages}{706--715}.
\newblock


\bibitem[\protect\citeauthoryear{Lei, Yu, Bansal, and Berg}{Lei
  et~al\mbox{.}}{2018}]%
        {lei2018tvqa}
\bibfield{author}{\bibinfo{person}{Jie Lei}, \bibinfo{person}{Licheng Yu},
  \bibinfo{person}{Mohit Bansal}, {and} \bibinfo{person}{Tamara~L Berg}.}
  \bibinfo{year}{2018}\natexlab{}.
\newblock \showarticletitle{TVQA: Localized, Compositional Video Question
  Answering}. In \bibinfo{booktitle}{\emph{EMNLP}}.
\newblock


\bibitem[\protect\citeauthoryear{Li, Chen, Cheng, Gan, Yu, and Liu}{Li
  et~al\mbox{.}}{2020}]%
        {li2020hero}
\bibfield{author}{\bibinfo{person}{Linjie Li}, \bibinfo{person}{Yen-Chun Chen},
  \bibinfo{person}{Yu Cheng}, \bibinfo{person}{Zhe Gan},
  \bibinfo{person}{Licheng Yu}, {and} \bibinfo{person}{Jingjing Liu}.}
  \bibinfo{year}{2020}\natexlab{}.
\newblock \showarticletitle{HERO: Hierarchical Encoder for Video+ Language
  Omni-representation Pre-training}.
\newblock \bibinfo{journal}{\emph{arXiv preprint arXiv:2005.00200}}
  (\bibinfo{year}{2020}).
\newblock


\bibitem[\protect\citeauthoryear{Lin, Zhao, and Shou}{Lin
  et~al\mbox{.}}{2017}]%
        {lin2017single}
\bibfield{author}{\bibinfo{person}{Tianwei Lin}, \bibinfo{person}{Xu Zhao},
  {and} \bibinfo{person}{Zheng Shou}.} \bibinfo{year}{2017}\natexlab{}.
\newblock \showarticletitle{Single shot temporal action detection}. In
  \bibinfo{booktitle}{\emph{Proceedings of the 25th ACM international
  conference on Multimedia}}. \bibinfo{pages}{988--996}.
\newblock


\bibitem[\protect\citeauthoryear{Liu, Chen, Cheng, Gan, Yu, Yang, and Liu}{Liu
  et~al\mbox{.}}{2020}]%
        {liu2020violin}
\bibfield{author}{\bibinfo{person}{Jingzhou Liu}, \bibinfo{person}{Wenhu Chen},
  \bibinfo{person}{Yu Cheng}, \bibinfo{person}{Zhe Gan},
  \bibinfo{person}{Licheng Yu}, \bibinfo{person}{Yiming Yang}, {and}
  \bibinfo{person}{Jingjing Liu}.} \bibinfo{year}{2020}\natexlab{}.
\newblock \showarticletitle{VIOLIN: A Large-Scale Dataset for
  Video-and-Language Inference}. In \bibinfo{booktitle}{\emph{CVPR 2020}}.
\newblock


\bibitem[\protect\citeauthoryear{Liu, Wang, Nie, He, Chen, and Chua}{Liu
  et~al\mbox{.}}{2018}]%
        {liu2018attentive}
\bibfield{author}{\bibinfo{person}{Meng Liu}, \bibinfo{person}{Xiang Wang},
  \bibinfo{person}{Liqiang Nie}, \bibinfo{person}{Xiangnan He},
  \bibinfo{person}{Baoquan Chen}, {and} \bibinfo{person}{Tat-Seng Chua}.}
  \bibinfo{year}{2018}\natexlab{}.
\newblock \showarticletitle{Attentive moment retrieval in videos}. In
  \bibinfo{booktitle}{\emph{The 41st International ACM SIGIR Conference on
  Research \& Development in Information Retrieval}}. ACM,
  \bibinfo{pages}{15--24}.
\newblock


\bibitem[\protect\citeauthoryear{Lu, Chen, Tan, Li, and Xiao}{Lu
  et~al\mbox{.}}{2019}]%
        {Lu2019DEBUGAD}
\bibfield{author}{\bibinfo{person}{Chujie Lu}, \bibinfo{person}{Long Chen},
  \bibinfo{person}{Chilie Tan}, \bibinfo{person}{Xiaolin Li}, {and}
  \bibinfo{person}{Jun Xiao}.} \bibinfo{year}{2019}\natexlab{}.
\newblock \showarticletitle{DEBUG: A Dense Bottom-Up Grounding Approach for
  Natural Language Video Localization}. In
  \bibinfo{booktitle}{\emph{EMNLP/IJCNLP}}.
\newblock


\bibitem[\protect\citeauthoryear{Luo and Shakhnarovich}{Luo and
  Shakhnarovich}{2017}]%
        {luo2017comprehension}
\bibfield{author}{\bibinfo{person}{Ruotian Luo} {and} \bibinfo{person}{Gregory
  Shakhnarovich}.} \bibinfo{year}{2017}\natexlab{}.
\newblock \showarticletitle{Comprehension-guided referring expressions}. In
  \bibinfo{booktitle}{\emph{Proceedings of the IEEE Conference on Computer
  Vision and Pattern Recognition}}. \bibinfo{pages}{7102--7111}.
\newblock


\bibitem[\protect\citeauthoryear{Mao, Huang, Toshev, Camburu, Yuille, and
  Murphy}{Mao et~al\mbox{.}}{2016}]%
        {mao2016generation}
\bibfield{author}{\bibinfo{person}{Junhua Mao}, \bibinfo{person}{Jonathan
  Huang}, \bibinfo{person}{Alexander Toshev}, \bibinfo{person}{Oana Camburu},
  \bibinfo{person}{Alan~L Yuille}, {and} \bibinfo{person}{Kevin Murphy}.}
  \bibinfo{year}{2016}\natexlab{}.
\newblock \showarticletitle{Generation and comprehension of unambiguous object
  descriptions}. In \bibinfo{booktitle}{\emph{Proceedings of the IEEE
  conference on computer vision and pattern recognition}}.
  \bibinfo{pages}{11--20}.
\newblock


\bibitem[\protect\citeauthoryear{Nagaraja, Morariu, and Davis}{Nagaraja
  et~al\mbox{.}}{2016}]%
        {nagaraja2016modeling}
\bibfield{author}{\bibinfo{person}{Varun~K Nagaraja}, \bibinfo{person}{Vlad~I
  Morariu}, {and} \bibinfo{person}{Larry~S Davis}.}
  \bibinfo{year}{2016}\natexlab{}.
\newblock \showarticletitle{Modeling context between objects for referring
  expression understanding}. In \bibinfo{booktitle}{\emph{European Conference
  on Computer Vision}}. Springer, \bibinfo{pages}{792--807}.
\newblock


\bibitem[\protect\citeauthoryear{Pennington, Socher, and Manning}{Pennington
  et~al\mbox{.}}{2014}]%
        {Pennington2014GloveGV}
\bibfield{author}{\bibinfo{person}{Jeffrey Pennington},
  \bibinfo{person}{Richard Socher}, {and} \bibinfo{person}{Christopher~D.
  Manning}.} \bibinfo{year}{2014}\natexlab{}.
\newblock \showarticletitle{Glove: Global Vectors for Word Representation}. In
  \bibinfo{booktitle}{\emph{Proceedings of the Conference on Empirical Methods
  in Natural Language Processing}}.
\newblock


\bibitem[\protect\citeauthoryear{Regneri, Rohrbach, Wetzel, Thater, Schiele,
  and Pinkal}{Regneri et~al\mbox{.}}{2013}]%
        {Regneri2013GroundingAD}
\bibfield{author}{\bibinfo{person}{Michaela Regneri}, \bibinfo{person}{Marcus
  Rohrbach}, \bibinfo{person}{Dominikus Wetzel}, \bibinfo{person}{Stefan
  Thater}, \bibinfo{person}{Bernt Schiele}, {and} \bibinfo{person}{Manfred
  Pinkal}.} \bibinfo{year}{2013}\natexlab{}.
\newblock \showarticletitle{Grounding Action Descriptions in Videos}.
\newblock \bibinfo{journal}{\emph{Transactions of the Association for
  Computational Linguistics}}  \bibinfo{volume}{1} (\bibinfo{year}{2013}),
  \bibinfo{pages}{25--36}.
\newblock


\bibitem[\protect\citeauthoryear{Rohrbach, Rohrbach, Hu, Darrell, and
  Schiele}{Rohrbach et~al\mbox{.}}{2016}]%
        {rohrbach2016grounding}
\bibfield{author}{\bibinfo{person}{Anna Rohrbach}, \bibinfo{person}{Marcus
  Rohrbach}, \bibinfo{person}{Ronghang Hu}, \bibinfo{person}{Trevor Darrell},
  {and} \bibinfo{person}{Bernt Schiele}.} \bibinfo{year}{2016}\natexlab{}.
\newblock \showarticletitle{Grounding of textual phrases in images by
  reconstruction}. In \bibinfo{booktitle}{\emph{European Conference on Computer
  Vision}}. Springer, \bibinfo{pages}{817--834}.
\newblock


\bibitem[\protect\citeauthoryear{Shou, Chan, Zareian, Miyazawa, and Chang}{Shou
  et~al\mbox{.}}{2017}]%
        {shou2017cdc}
\bibfield{author}{\bibinfo{person}{Zheng Shou}, \bibinfo{person}{Jonathan
  Chan}, \bibinfo{person}{Alireza Zareian}, \bibinfo{person}{Kazuyuki
  Miyazawa}, {and} \bibinfo{person}{Shih-Fu Chang}.}
  \bibinfo{year}{2017}\natexlab{}.
\newblock \showarticletitle{Cdc: Convolutional-de-convolutional networks for
  precise temporal action localization in untrimmed videos}. In
  \bibinfo{booktitle}{\emph{Proceedings of the IEEE conference on computer
  vision and pattern recognition}}. \bibinfo{pages}{5734--5743}.
\newblock


\bibitem[\protect\citeauthoryear{Shou, Wang, and Chang}{Shou
  et~al\mbox{.}}{2016}]%
        {shou2016temporal}
\bibfield{author}{\bibinfo{person}{Zheng Shou}, \bibinfo{person}{Dongang Wang},
  {and} \bibinfo{person}{Shih-Fu Chang}.} \bibinfo{year}{2016}\natexlab{}.
\newblock \showarticletitle{Temporal action localization in untrimmed videos
  via multi-stage cnns}. In \bibinfo{booktitle}{\emph{Proceedings of the IEEE
  Conference on Computer Vision and Pattern Recognition}}.
  \bibinfo{pages}{1049--1058}.
\newblock


\bibitem[\protect\citeauthoryear{Tran, Bourdev, Fergus, Torresani, and
  Paluri}{Tran et~al\mbox{.}}{2014}]%
        {Tran2014LearningSF}
\bibfield{author}{\bibinfo{person}{Du Tran}, \bibinfo{person}{Lubomir~D.
  Bourdev}, \bibinfo{person}{Rob Fergus}, \bibinfo{person}{Lorenzo Torresani},
  {and} \bibinfo{person}{Manohar Paluri}.} \bibinfo{year}{2014}\natexlab{}.
\newblock \showarticletitle{Learning Spatiotemporal Features with 3D
  Convolutional Networks}. In \bibinfo{booktitle}{\emph{Proceedings of the IEEE
  International Conference on Computer Vision (ICCV)}}.
  \bibinfo{pages}{4489--4497}.
\newblock


\bibitem[\protect\citeauthoryear{Vaswani, Shazeer, Parmar, Uszkoreit, Jones,
  Gomez, Kaiser, and Polosukhin}{Vaswani et~al\mbox{.}}{2017}]%
        {Vaswani2017AttentionIA}
\bibfield{author}{\bibinfo{person}{Ashish Vaswani}, \bibinfo{person}{Noam
  Shazeer}, \bibinfo{person}{Niki Parmar}, \bibinfo{person}{Jakob Uszkoreit},
  \bibinfo{person}{Llion Jones}, \bibinfo{person}{Aidan~N. Gomez},
  \bibinfo{person}{Lukasz Kaiser}, {and} \bibinfo{person}{Illia Polosukhin}.}
  \bibinfo{year}{2017}\natexlab{}.
\newblock \showarticletitle{Attention is All you Need}. In
  \bibinfo{booktitle}{\emph{Proceedings of the Thirty-third Conference on
  Neural Information Processing Systems (NeurIPS)}}.
\newblock


\bibitem[\protect\citeauthoryear{Wang, Cheng, and Schmidt~Feris}{Wang
  et~al\mbox{.}}{2016a}]%
        {Wang2016CVPR}
\bibfield{author}{\bibinfo{person}{Jing Wang}, \bibinfo{person}{Yu Cheng},
  {and} \bibinfo{person}{Rogerio Schmidt~Feris}.}
  \bibinfo{year}{2016}\natexlab{a}.
\newblock \showarticletitle{Walk and Learn: Facial Attribute Representation
  Learning From Egocentric Video and Contextual Data}. In
  \bibinfo{booktitle}{\emph{CVPR 2016}}.
\newblock


\bibitem[\protect\citeauthoryear{Wang, Li, and Lazebnik}{Wang
  et~al\mbox{.}}{2016b}]%
        {wang2016learning}
\bibfield{author}{\bibinfo{person}{Liwei Wang}, \bibinfo{person}{Yin Li}, {and}
  \bibinfo{person}{Svetlana Lazebnik}.} \bibinfo{year}{2016}\natexlab{b}.
\newblock \showarticletitle{Learning deep structure-preserving image-text
  embeddings}. In \bibinfo{booktitle}{\emph{Proceedings of the IEEE conference
  on computer vision and pattern recognition}}. \bibinfo{pages}{5005--5013}.
\newblock


\bibitem[\protect\citeauthoryear{Xu, Das, and Saenko}{Xu et~al\mbox{.}}{2017}]%
        {xu2017r}
\bibfield{author}{\bibinfo{person}{Huijuan Xu}, \bibinfo{person}{Abir Das},
  {and} \bibinfo{person}{Kate Saenko}.} \bibinfo{year}{2017}\natexlab{}.
\newblock \showarticletitle{R-c3d: Region convolutional 3d network for temporal
  activity detection}. In \bibinfo{booktitle}{\emph{Proceedings of the IEEE
  international conference on computer vision}}. \bibinfo{pages}{5783--5792}.
\newblock


\bibitem[\protect\citeauthoryear{Xu, He, Plummer, Sigal, Sclaroff, and
  Saenko}{Xu et~al\mbox{.}}{2019}]%
        {xu2019multilevel}
\bibfield{author}{\bibinfo{person}{Huijuan Xu}, \bibinfo{person}{Kun He},
  \bibinfo{person}{Bryan~A Plummer}, \bibinfo{person}{Leonid Sigal},
  \bibinfo{person}{Stan Sclaroff}, {and} \bibinfo{person}{Kate Saenko}.}
  \bibinfo{year}{2019}\natexlab{}.
\newblock \showarticletitle{Multilevel language and vision integration for
  text-to-clip retrieval}. In \bibinfo{booktitle}{\emph{Proceedings of the AAAI
  Conference on Artificial Intelligence}}, Vol.~\bibinfo{volume}{33}.
  \bibinfo{pages}{9062--9069}.
\newblock


\bibitem[\protect\citeauthoryear{Yeung, Russakovsky, Mori, and Fei-Fei}{Yeung
  et~al\mbox{.}}{2016}]%
        {yeung2016end}
\bibfield{author}{\bibinfo{person}{Serena Yeung}, \bibinfo{person}{Olga
  Russakovsky}, \bibinfo{person}{Greg Mori}, {and} \bibinfo{person}{Li
  Fei-Fei}.} \bibinfo{year}{2016}\natexlab{}.
\newblock \showarticletitle{End-to-end learning of action detection from frame
  glimpses in videos}. In \bibinfo{booktitle}{\emph{Proceedings of the IEEE
  Conference on Computer Vision and Pattern Recognition}}.
  \bibinfo{pages}{2678--2687}.
\newblock


\bibitem[\protect\citeauthoryear{Yu, Lin, Shen, Yang, Lu, Bansal, and Berg}{Yu
  et~al\mbox{.}}{2018}]%
        {yu2018mattnet}
\bibfield{author}{\bibinfo{person}{Licheng Yu}, \bibinfo{person}{Zhe Lin},
  \bibinfo{person}{Xiaohui Shen}, \bibinfo{person}{Jimei Yang},
  \bibinfo{person}{Xin Lu}, \bibinfo{person}{Mohit Bansal}, {and}
  \bibinfo{person}{Tamara~L Berg}.} \bibinfo{year}{2018}\natexlab{}.
\newblock \showarticletitle{Mattnet: Modular attention network for referring
  expression comprehension}. In \bibinfo{booktitle}{\emph{Proceedings of the
  IEEE Conference on Computer Vision and Pattern Recognition}}.
  \bibinfo{pages}{1307--1315}.
\newblock


\bibitem[\protect\citeauthoryear{Yuan, Ma, Wang, Liu, and Zhu}{Yuan
  et~al\mbox{.}}{2019a}]%
        {SCDM19}
\bibfield{author}{\bibinfo{person}{Yitian Yuan}, \bibinfo{person}{Lin Ma},
  \bibinfo{person}{Jingwen Wang}, \bibinfo{person}{Wei Liu}, {and}
  \bibinfo{person}{Wenwu Zhu}.} \bibinfo{year}{2019}\natexlab{a}.
\newblock \showarticletitle{Semantic Conditioned Dynamic Modulation for
  Temporal Sentence Grounding in Videos}. In
  \bibinfo{booktitle}{\emph{Proceedings of the Thirty-third Conference on
  Neural Information Processing Systems (NeurIPS)}}.
\newblock


\bibitem[\protect\citeauthoryear{Yuan, Mei, and Zhu}{Yuan
  et~al\mbox{.}}{2019b}]%
        {yuan2019find}
\bibfield{author}{\bibinfo{person}{Yitian Yuan}, \bibinfo{person}{Tao Mei},
  {and} \bibinfo{person}{Wenwu Zhu}.} \bibinfo{year}{2019}\natexlab{b}.
\newblock \showarticletitle{To find where you talk: Temporal sentence
  localization in video with attention based location regression}. In
  \bibinfo{booktitle}{\emph{Proceedings of the AAAI Conference on Artificial
  Intelligence}}, Vol.~\bibinfo{volume}{33}. \bibinfo{pages}{9159--9166}.
\newblock


\bibitem[\protect\citeauthoryear{Zeng, Gan, Chen, Huang, Wu, and Tan}{Zeng
  et~al\mbox{.}}{2019}]%
        {zeng2019breaking}
\bibfield{author}{\bibinfo{person}{Runhao Zeng}, \bibinfo{person}{Chuang Gan},
  \bibinfo{person}{Peihao Chen}, \bibinfo{person}{Wenbing Huang},
  \bibinfo{person}{Qingyao Wu}, {and} \bibinfo{person}{Mingkui Tan}.}
  \bibinfo{year}{2019}\natexlab{}.
\newblock \showarticletitle{Breaking winner-takes-all: Iterative-winners-out
  networks for weakly supervised temporal action localization}.
\newblock \bibinfo{journal}{\emph{IEEE Transactions on Image Processing}}
  \bibinfo{volume}{28}, \bibinfo{number}{12} (\bibinfo{year}{2019}),
  \bibinfo{pages}{5797--5808}.
\newblock


\bibitem[\protect\citeauthoryear{Zhang, Dai, Wang, Wang, and Davis}{Zhang
  et~al\mbox{.}}{2019a}]%
        {Zhang_2019_CVPR}
\bibfield{author}{\bibinfo{person}{Da Zhang}, \bibinfo{person}{Xiyang Dai},
  \bibinfo{person}{Xin Wang}, \bibinfo{person}{Yuan-Fang Wang}, {and}
  \bibinfo{person}{Larry~S. Davis}.} \bibinfo{year}{2019}\natexlab{a}.
\newblock \showarticletitle{MAN: Moment Alignment Network for Natural Language
  Moment Retrieval via Iterative Graph Adjustment}. In
  \bibinfo{booktitle}{\emph{Proceedings of the IEEE Conference on Computer
  Vision and Pattern Recognition}}.
\newblock


\bibitem[\protect\citeauthoryear{Zhang, Niu, and Chang}{Zhang
  et~al\mbox{.}}{2018}]%
        {zhang2018grounding}
\bibfield{author}{\bibinfo{person}{Hanwang Zhang}, \bibinfo{person}{Yulei Niu},
  {and} \bibinfo{person}{Shih-Fu Chang}.} \bibinfo{year}{2018}\natexlab{}.
\newblock \showarticletitle{Grounding referring expressions in images by
  variational context}. In \bibinfo{booktitle}{\emph{Proceedings of the IEEE
  Conference on Computer Vision and Pattern Recognition}}.
  \bibinfo{pages}{4158--4166}.
\newblock


\bibitem[\protect\citeauthoryear{Zhang, Lin, Zhao, and Xiao}{Zhang
  et~al\mbox{.}}{2019b}]%
        {Zhang2019CrossModalIN}
\bibfield{author}{\bibinfo{person}{Zhu Zhang}, \bibinfo{person}{Zhijie Lin},
  \bibinfo{person}{Zhou Zhao}, {and} \bibinfo{person}{Zhenxin Xiao}.}
  \bibinfo{year}{2019}\natexlab{b}.
\newblock \showarticletitle{Cross-Modal Interaction Networks for Query-Based
  Moment Retrieval in Videos}. In \bibinfo{booktitle}{\emph{The 42nd
  International ACM SIGIR Conference on Research and Development in
  Information}}. ACM.
\newblock


\bibitem[\protect\citeauthoryear{Zhao, Xiong, Wang, Wu, Tang, and Lin}{Zhao
  et~al\mbox{.}}{2017}]%
        {zhao2017temporal}
\bibfield{author}{\bibinfo{person}{Yue Zhao}, \bibinfo{person}{Yuanjun Xiong},
  \bibinfo{person}{Limin Wang}, \bibinfo{person}{Zhirong Wu},
  \bibinfo{person}{Xiaoou Tang}, {and} \bibinfo{person}{Dahua Lin}.}
  \bibinfo{year}{2017}\natexlab{}.
\newblock \showarticletitle{Temporal action detection with structured segment
  networks}. In \bibinfo{booktitle}{\emph{Proceedings of the IEEE International
  Conference on Computer Vision}}. \bibinfo{pages}{2914--2923}.
\newblock


\end{thebibliography}
\end{document}